\documentclass[12pt]{spieman}  
\usepackage{amsmath,amsfonts,amssymb}
\usepackage{graphicx}
\usepackage{setspace}
\usepackage{tocloft}
\usepackage{array}
\usepackage{mathabx}
\usepackage{url}

\def\D{{\mathcal{D}}}
\def\W{{\mathcal{W}}}
\def\F{{\mathcal{F}}}
\def\x{{\mathbf{x}}}

\title{Wavelet Burst Accumulation for turbulence mitigation.}

\author[a]{J\'er\^ome Gilles}
\author[b]{Stanley Osher}
\affil[a]{Department of Mathematics and Statistics, San Diego State University, San Diego, CA, 92182-7720 USA}
\affil[b]{Department of Mathematics, University of California Los Angeles, Los Angeles, CA, 90095 USA}

\cftpagenumbersoff{figure}
\cftpagenumbersoff{table} 
\begin{document} 
\maketitle

\begin{abstract}
In this paper, we investigate the extension of the recently proposed weighted Fourier burst accumulation (FBA) method into the wavelet domain. The purpose of FBA is to 
reconstruct a clean and sharp image from a sequence of blurred frames. This concept lies in the construction of weights to amplify dominant frequencies in the Fourier spectrum of 
each frame. The 
reconstructed image is then obtained by taking the inverse Fourier transform of the average of all processed spectra. In this paper, we first suggest to replace the rigid 
registration step used in the 
original algorithm by a non-rigid registration in order to be able to process sequences acquired through atmospheric turbulence. Second, we propose to work in a wavelet domain 
instead of the Fourier one. This leads us to the construction of two types of algorithms. Finally, we propose an alternative approach to replace the 
weighting idea by an approach promoting the sparsity in the used space. Several experiments are provided to illustrate the efficiency of the proposed methods.
\end{abstract}


{\noindent \footnotesize\textbf{*} Corresponding author: J\'er\^ome Gilles, \linkable{jgilles@mail.sdsu.edu} }


\section{Introduction}
In the last decade, turbulence image mitigation algorithms gained a lot of interest since long range imaging systems have been developed to improve target 
identification. Such problem is challenging because it corresponds to the propagation of light through a random media (i.e the atmospheric optical properties are not homogeneous) resulting in two main effects on the observed images: 
geometric distortions and blur. Such situation can occur in many scenarios: for example, underwater imaging which is subject to the water scattering effects, or video shooting during the summer which suffers from hot air turbulence near the ground, and so on. Weak and medium turbulence do not really affect human observers, but they can be critical for automatic target recognition algorithms. Indeed, shapes of objects may be very different from those learned by the algorithm.\\
Local filtering techniques \cite{Lemaitre:2007p7023} (Wiener filter, Laplacian regularization, \ldots) were first proposed in the literature to retrieve a ``clear image''. Such methods are implemented by block partitioning of the image, having the main issue that they result in some block artifacts on the restored images. In \cite{Frakes:2001p7017,Gepshtein:2004p7030}, the authors proposed to model the turbulence phenomenon by using two operators:
\begin{equation}\label{eq:turbulence}
v_i(\x)=D_i(H(u(\x)))+\text{noise},
\end{equation}
where $u$ is the static original scene we want to retrieve, $\x$ the pixel location, $v_i$ the observed image at time $i$, $H$ a blurring kernel, and $D_i$ represents the geometric distortions at time $i$ (it is commonly accepted that the blur is a stationary phenomenon compared to geometric distortions). Based on this model, in \cite{Gilles:2008p4901}, and then improved by combining \cite{MaoGilles} and \cite{Gilles2012}, the authors proposed methods to evaluate the inverse operators $H^{-1}$ and $D_i^{-1}$. In \cite{Mario1}, the authors took a system 
point of view. They combined a Kalman filter to stabilize the deformations with a Nonlocal Total Variation \cite{Gilboa2008} deconvolution step. A centroid based approach is 
proposed in \cite{Mario2} and \cite{Mario3}.  
In \cite{Li2007}, assuming long exposure video capture, the authors used Principal Component Analysis to find the statistically best restored image. In 
\cite{Aubailly2009}, the authors followed the assumption that for a given location in the image, at some specific timings, its neighborhood has some high probability to appear with good quality (this principle is called the ``\textit{lucky-region}'' approach). The restored region is then a fusion of the ``best'' ones. A spatially variant deblurring approach was 
proposed in \cite{Hirsch2010} but it does not address the geometrical distortion issue. Based on Frake's model described above, in \cite{Zhu2010} the authors used a B-Spline 
registration algorithm embedded in a Bayesian framework with bilateral total variation (TV) regularization to invert the geometric distortions and the blur. In \cite{Lou2013}, the 
authors proposed to obtain a latent (stabilized) image from the sequence $\{v_i\}$ by combining Sobolev gradients and Laplacian in a unified framework. In \cite{Zwart}, a precise 
and rapid control grid optical flow estimation algorithm is proposed and used to remove geometric distortions due to turbulence. In \cite{Halder,Halder2}, the authors designed a two 
steps method: first they used a multiscale optical flow estimation and second, they introduced an algorithm, called \textit
{First Register Then Average And Subtract}, to reconstruct 
a restored image. In \cite{Song}, the SURE (Stein's unbiased risk estimate) method is used to optimize an objective function, which combines temporal and spatial information, to obtain a restored image. An 
algorithm combining non-rigid registration and the lucky-region approach is proposed in \cite{Yang}. In \cite{Halder3}, a generalized regression neural network technique is used to 
learn, predict and then compensate turbulence induced deformations. In \cite{Anantrasirichai2013}, the authors proposed a method fusing coefficients from a dual tree complex wavelet domain. The fusion technique is based on the selection of informative regions of interest (ROI) and using a non-rigid registration technique. The proposed approach has two main drawbacks: it requires a manual selection of the used ROIs in the image and it uses a segmentation step in the fusion strategy. Such segmentation step can be rapidly limiting, indeed,  the ``general'' segmentation problem is still an open problem and results are highly dependent on the type of images (textured v.s non-textures for instance).\\
Recently, outside the context of turbulence degradations, the authors of \cite{Delbracio2015} (and extended in \cite{Delbracio2015a} to videos with moving objects) proposed a simple algorithm to remove motion blur from a sequence a observation. This technique consists in a weighted accumulation of all frames in the Fourier space. In this paper, we investigate the opportunity to extend such accumulation process to restore images acquired through atmospheric turbulence. First, we propose to incorporate a non-rigid registration step in the algorithm given in \cite{Delbracio2015}. Second, we generalize 
the concept of weighted accumulation in wavelet spaces instead of the Fourier space. We design two different algorithms and study their mathematical properties. We also 
suggest an alternative to the weighting process based on a sparsity constraint. Several 
experiments ran on real data with different wavelet families are presented to illustrate advantages of using wavelets instead of Fourier. The paper is 
organized as follows: Section~\ref{sec:FBA} recalls the weighted Fourier burst accumulation algorithm proposed in \cite{Delbracio2015}. In section~\ref{sec:nr}, we propose the use of an 
initial non-rigid registration step. Section~\ref{sec:wavburst} introduces the generalization to the wavelet domain while in Section~\ref{sec:math} we investigate some mathematical 
properties. In Section~\ref{sec:sba}, we introduce the concept of sparse burst accumulation where a sparsity constraint is used instead of the weighting method. 
Section~\ref{sec:expe} presents experimental results. Finally, we conclude the paper in Section~\ref{sec:conc}.

\section{Fourier burst accumulation}\label{sec:FBA}
In this section we recall the concept of weighted Fourier Burst Accumulation (FBA) proposed in \cite{Delbracio2015} and we introduce some notations which will be used throughout 
this paper. The aim of the FBA method is to retrieve a deblurred image of an original scene, denoted $u$, from a sequence of observations, assuming that those observations are 
affected by hand-shake blur. We will denote $\{v_i\}_{i=1}^M$ the sequence of $M$ observations such that, $\forall i=1,\ldots,M$, 
\begin{equation}
v_i(\x)=(k_i\ast u)(\x)+n_i(\x),
\end{equation}
where $\ast$ denotes the convolution product, $\x$ the position in the image, $n_i$ some noise and $k_i$ is a blurring kernel affecting $u$ at the frame $i$. \\
The authors of \cite{Delbracio2015} defined the weighted Fourier Burst Accumulation (FBA) algorithm in the following way: let $p$ be a positive integer, the restored image, $u_p$, 
is obtained by 
\begin{gather}
u_p(\x)=FBA(\{v_i(\x)\}_{i=1}^M\},p)=\F^{-1}\left(\sum_{i=1}^M w_i(\xi)\F(v_i)(\xi)\right)(\x),\\ \notag
\text{with}\qquad w_i(\xi)=\frac{G_\sigma(|\F(v_i)(\xi)|^p)}{\sum_{j=1}^M G_\sigma(|\F(v_j)(\xi)|^p)}.
\end{gather}

Where $\F$ denotes the Fourier transform ($\xi$ are the frequencies) and $G_\sigma$ is a Gaussian filtering of standard deviation $\sigma$. The authors mention that the algorithm 
is not sensitive to the 
choice of $\sigma$ and it can be automatically set to $\sigma=\min(m_w,m_h)/50$ (where $m_w$ and $m_h$ are respectively the width and height of the image). In practice, the 
authors apply first a preprocessing registration step (they use a combination of SIFT \cite{Lowe1999} and ORSA \cite{Fisher} algorithms) in order to remove affine 
transformations (translations, rotations and homothety). The authors also suggest that optional denoising and sharpening final steps can be applied for contrast enhancement 
purposes.\\ 
This approach is based on the following principle: Fourier coefficients which represent an important information in the image must appear consistently in all observations. 
Therefore, the weights $w_i(\xi)$ are large if Fourier coefficients $\F(v_i)(\xi)$ are relatively stable through time, and weak otherwise. Then multiplying the Fourier spectrum of 
each observation by these weights will amplify the important information and attenuate time non-consistent information. Finally, an average spectrum is computed and $u_p$ can be easily 
retrieved by inverse Fourier transform.

\section{Non-rigid regularization}\label{sec:nr}
As mentioned in the previous section, the authors of \cite{Delbracio2015} use an initial rigid registration step before applying the FBA algorithm. Unfortunately, for atmospheric turbulence mitigation 
purposes the observations are affected by non-rigid deformation and the use of a registration step based on SIFT and ORSA algorithms is inefficient. 
In \cite{MaoGilles}, a non-rigid regularization technique was proposed to remove atmospheric non-rigid distortions. It consists in solving the following variational problem
\begin{gather}\label{BasicOptimization}
(\hat{u},\{\hat{\Phi}_i\})=\arg_{u,\Phi_i}\min J(u)\quad \text{s.t.}\quad \forall i,v_i=\Phi_i u+\text{noise}, 
\end{gather}
where each $\Phi_i$ corresponds to geometric distortions induced by the turbulence on the original scene $u$ at time $i$. The term $J(u)$ is a regularizer permitting to 
introduce some constraints on the expected restored image $\hat{u}$. In \cite{MaoGilles}, the authors proposed to use the Non-Local Total Variation (NLTV) but other regularizers like 
the Total Variation or sparsity in a wavelet-type representation can also be used. The deformation fields $\Phi_i$ can be estimated via some optical-flow algorithm. Thus, the 
algorithm consists in iterating two steps: a non-rigid registration step and a regularization step. In this paper, we only keep the non-rigid registration step. In details, it 
consists first to compute an average frame $v_a(\x)=\frac{1}{M}\sum_{i=1}^M v_i(\x)$. Next, the deformation mappings $\Phi_i$, such that $\forall i=1,\ldots,M; v_i=\Phi_i(v_a)$, 
are estimated via an optical flow algorithm (in this paper, we use the multiscale Lukas-Kanade algorithm \cite{Bouguet}). Finally, the inverse mappings based on bilinear interpolations are applied to each frame: 
$\Phi_i^{-1}(v_i)\to v_i$. This registered sequence can then be used as the input of the burst accumulation methods.

\section{Weighted wavelet burst accumulation}\label{sec:wavburst}
Other image representations than the Fourier one are widely used in image processing, notably wavelet type representations (wavelets, framelets, curvelets,\ldots). Given a 
family of $N$ wavelets $\{\psi_n\}_{n=1}^N$, the wavelet representation of an image $v$ is given by the set of projections $\{v^n\}_{n=1}^N=\{\langle 
v,\psi_n\rangle\}_{n=1}^N$. In this paper, we will denote $\W$ a wavelet type transform of the image $v$, i.e $\{v^n\}_{n=1}^N=\W(v)$. The inverse wavelet transform will be 
denoted 
$\W^{-1}$, i.e $v=\W^{-1}(\{v^n\}_{n=1}^N)$.\\
We can define two types of burst accumulation approaches. The first one processes the burst accumulation directly in the wavelet domain, we will denote it WWBA (Weighted Wavelet 
Burst Accumulation). Here, the weights are computed from the wavelet coefficients themselves, in each subband, amplifying the most dominant coefficients through time. The 
corresponding formulation is given by equations~\eqref{eq:wwba}. 
\begin{gather}\label{eq:wwba}
u_p(\x)=WWBA(\{v_i\},\{\psi_n\},p)=\W^{-1}\left(\left\{\sum_{i=1}^M w_i^n(\x)v_i^n(\x)\right\}_{n=1}^N\right),\\
\text{with}\qquad w_i^n(\x)=\frac{G_\sigma(|v_i^n(\x)|^p)}{\sum_{j=1}^M G_\sigma(|v_j^n(\x)|^p)}. \notag
\end{gather}
The Gaussian filter, $G_\sigma$, is chosen as in the Fourier case.\\ 
The second type of accumulation consists in doing first the wavelet decomposition of each input frame. Then we apply the 
Fourier Burst Accumulation algorithm from \cite{Delbracio2015} (recalled in Section~\ref{sec:FBA}) to each subband sequences $\{v_i^n\}_{i=1}^M$. We call this approach the Weighted 
Wavelet Fourier Burst Accumulation (WWFBA) and can be formulated by equations~\eqref{eq:wwfba}.
\begin{equation}\label{eq:wwfba}
u_p(\x)=WWFBA(\{v_i\},\{\psi_n\},p)=\W^{-1}\left(\left\{FBA(\{v_i^n(\x)\}_{i=1}^M,p)\right\}_{n=1}^N\right).
\end{equation}
This option is equivalent to deblur the wavelet coefficients themselves.

\section{Mathematical characterization}\label{sec:math}
In \cite{Delbracio2015}, the authors characterized the FBA algorithm with the following proposition.\\
\vspace{2ex}\noindent{\footnotesize\textbf{Proposition 1.}
Applying the FBA algorithm on a sequence $\{v_i\}_{i=1}^M$ provides a restored image $u_p$ equivalent to 
\begin{equation}
u_p=k_e\ast u+\bar{n} \qquad\text{where} \qquad k_e=\F^{-1}\left(\sum_{i=1}^M w_i\F(k_i)\right),
\end{equation}
and $\bar{n}$ is the weighted average of the input noise.\\}
In this section, we investigate the opportunity to get the same kind of characterization for the two previously introduced algorithms in the wavelet domain described by \eqref{eq:wwba} and \eqref{eq:wwfba} in the previous section. First, we need to switch to the filter bank point of view about the wavelet decomposition. In the previous section, we denoted the wavelet decomposition of $v$ by $\{v^n\}_{n=1}^N=\{\langle v,\psi_n\rangle\}_{n=1}^N$. It is also well-known in wavelet theory, using the frame 
formalism \cite{Christensen2001}, that it can be written as a convolution product: 
\begin{equation}
\{v^n\}_{n=1}^N=\{v\ast \overline{\psi}_n\}_{n=1}^N,
\end{equation}
where $\overline{\psi}_n(\x)=\psi_n(-\x)$. The inverse wavelet transform is given by 
\begin{equation}
v(\x)=\W^{-1}(\{v^n\}_{n=1}^N)(\x)=\sum_{n=1}^N (v^n\ast\psi_n)(\x).
\end{equation}
Therefore, in the same formalism, the wavelet transform of each observed frame $v_i$ can be written as (we omit $\x$ in the following in order to simplify the notations) 
\begin{equation}
\{v_i^n\}_{n=1}^N=\{v_i\ast \overline{\psi}_n\}_{n=1}^N=\{(k_i\ast u+n_i)\ast\overline{\psi}_n\}_{n=1}^N\\=\{k_i\ast u\ast \overline{\psi}_n\}_{n=1}^N+\{n_i\ast \overline{\psi}_n\}_{n=1}^N.
\end{equation}
\subsection{Analysis of the WWBA algorithm}
The WWBA restoration algorithm proposed in the previous section is equivalent to 
\begin{align}
u_p&=\W^{-1}\left(\left\{\sum_{i=1}^M w_i^nv_i^n\right\}_{n=1}^N\right)\\
&=\sum_{n=1}^N\left(\left(\sum_{i=1}^M w_i^nv_i^n\right)\ast\psi_n\right)\\
&=\sum_{n=1}^N\sum_{i=1}^M \left(w_i^n(k_i\ast u\ast \overline{\psi}_n+n_i\ast \overline{\psi}_n)\right)\ast\psi_n\\
&=\sum_{i=1}^M\sum_{n=1}^N (w_i^n(k_i\ast u\ast \overline{\psi}_n))\ast\psi_n+\sum_{i=1}^M\sum_{n=1}^N (w_i^n(n_i\ast \overline{\psi}_n))\ast\psi_n.
\end{align}
Unfortunately, pointwise multiplication and convolution do not commute so it is not possible to write this expression as the convolution of $u$ with some equivalent kernel as the authors did in \cite{Delbracio2015} for the Fourier burst case. 

\subsection{Analysis of the WWFBA algorithm}
In the WWFBA algorithm case, we have the following proposition.

\vspace{2ex}\noindent{\footnotesize\textbf{Proposition 2.}
Applying independent Fourier burst accumulations on each subband ($n$) sequences $\{v_i^n\}_{i=1}^{M}$ is equivalent to 
\begin{equation}
u_p=(k_e\ast u)+\overline{n}\qquad \text{where}\qquad k_e=\sum_{n=1}^N\sum_{i=1}^M \F^{-1}(w_i^n)\ast k_i\ast \overline{\psi}_n\ast\psi_n.
\end{equation}}
\noindent\underline{Proof}: In the following, we use indistinctly $\F(u)$ or $\hat{u}$ to designate the Fourier transform and $\F^{-1}(u)$ or $\check{u}$ for the inverse transform in order to simplify the notations.
\begin{align}
u_p&=\W^{-1}\left(\left\{FBA\left(\{v_i^n\}_{i=1}^M,p\right)\right\}_{n=1}^N\right)\\
&=\W^{-1}\left(\left\{\F^{-1}\left(\sum_{i=1}^M w_i^n\F(v_i^n)\right)\right\}_{n=1}^N\right)\\
&=\W^{-1}\left(\left\{\F^{-1}\left(\sum_{i=1}^M w_i^n\left(\hat{k_i}\hat{u}\widehat{\overline{\psi}_n}+\hat{n_i}\widehat{\overline{\psi}_n}\right)\right)\right\}_{n=1}^N\right)\\
&=\W^{-1}\left(\left\{\sum_{i=1}^M \F^{-1}\left(w_i^n\hat{k_i}\hat{u}\widehat{\overline{\psi}_n}+w_i^n\hat{n_i}\widehat{\overline{\psi}_n}\right)\right\}_{n=1}^N\right)\\
&=\W^{-1}\left(\left\{\sum_{i=1}^M \widecheck{w_i^n}\ast k_i\ast u\ast \overline{\psi}_n+\widecheck{w_i^n}\ast n_i \ast \overline{\psi}_n\right\}_{n=1}^N\right).
\end{align}
Using the expression of the inverse wavelet transform, we get
\begin{align}
u_p&=\sum_{n=1}^N\left(\sum_{i=1}^M \widecheck{w_i^n}\ast k_i\ast u\ast \overline{\psi}_n+\widecheck{w_i^n}\ast n_i \ast \overline{\psi}_n\right)\ast\psi_n\\
&=\sum_{n=1}^N\sum_{i=1}^M \left(\widecheck{w_i^n}\ast k_i\ast u\ast \overline{\psi}_n\ast\psi_n+\widecheck{w_i^n}\ast n_i \ast \overline{\psi}_n\ast\psi_n\right)\\
&=\sum_{i=1}^M\sum_{n=1}^N \widecheck{w_i^n}\ast k_i\ast \overline{\psi}_n\ast\psi_n\ast u+\sum_{i=1}^M\sum_{n=1}^N \widecheck{w_i^n}\ast n_i\ast \overline{\psi}_n\ast\psi_n.
\end{align}
If we denote 
\begin{equation}
\overline{n}=\sum_{i=1}^M\sum_{n=1}^N \widecheck{w_i^n}\ast n_i\ast \overline{\psi}_n\ast\psi_n,
\end{equation}
and
\begin{equation}
k_e=\sum_{i=1}^M\sum_{n=1}^N \widecheck{w_i^n}\ast k_i\ast \overline{\psi}_n\ast\psi_n,
\end{equation}
then the wavelet burst accumulation is equivalent to $u_p=k_e\ast u+\overline{n}$ which ends the proof.

\section{Non-linear burst accumulation}\label{sec:sba}
The frame accumulation techniques described in section \ref{sec:wavburst} are all based  on linear combination (either by pointwise multiplications or convolutions) of each frames in 
different representation domains. These weights basically correspond to amplify dominant coefficients thus we can ask if it is possible to exploit other approaches to perform such 
amplification? In this section, we propose an alternative to the use of weights. Amplifying only the dominant coefficients can be interpreted in the sense that only those 
dominant coefficients are important to represent the image. Therefore, instead of using some amplification, we can imagine to keep the dominant coefficients and remove the other ones. In other words the restored image is expected to have a sparse representation in the used representation domains. It is then 
natural to promote the sparsity in each representation before doing the accumulation. In the last decade, the compressive sensing community widely developed such concepts. It 
is notably well established that minimizing models based on $L^1$-norm provide sparse solutions \cite{Goldstein2009}. Let a function $f$, the simplest model to find a sparse representation $g$ from $f$ is 
given by \eqref{eq:l1}.
\begin{equation}\label{eq:l1}
g=\underset{{\tilde{g}}}{\arg}\min \|\tilde{g}\|_1+\frac{1}{2\lambda}\|\tilde{g}-f\|_2^2.
\end{equation}
It is well-known that the solution of this minimization problem is given by soft-thresholding $f$ with threshold $\lambda$, and is given by (the operators are understood pointwise)
\begin{equation}
g=Soft(f,\lambda)=\frac{f}{|f|}\max(|f|-\lambda,0).
\end{equation}
Therefore, denoting $\D$ the representation domain which can be either $\F$ or $\W$, we propose the following general sparse burst accumulation model.
\begin{equation}\label{eq:sba}
u_p(\x)=S\D BA(\{v_i(\x)\}_{i=1}^M\},\lambda)=\D^{-1}\left(\sum_{i=1}^M Soft(\D(v_i),\lambda)\right)(\x).
\end{equation}
Notice that in the case of a wavelet-type decomposition, the soft-thresholding operator is applied in each subband.

\section{Experiments}\label{sec:expe}
\begin{figure}[!t]
\begin{center}
\includegraphics[width=0.6\textwidth]{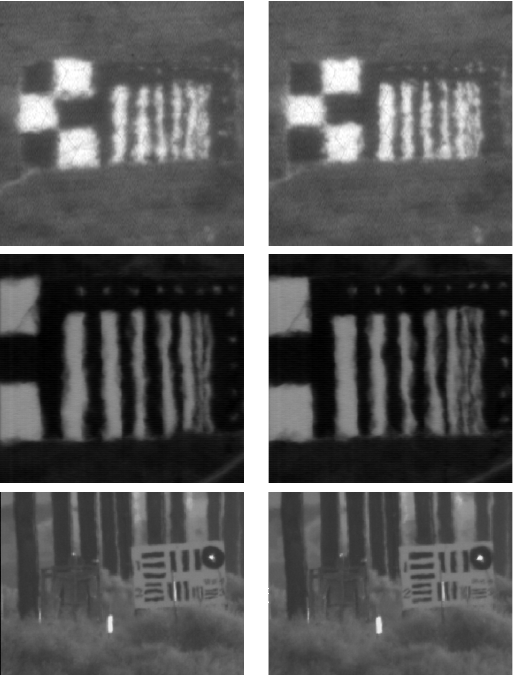} 
\end{center}
\caption{Original sequences: Barchart 1 (top row), Barchart 2 (middle row), Barchart 3 (bottom row). The consecutive columns correspond to the frames 1 and 25, respectively.}
\label{fig:barchart}
\end{figure}

\begin{figure}[!t]
\begin{center}
\includegraphics[height=0.94\textheight]{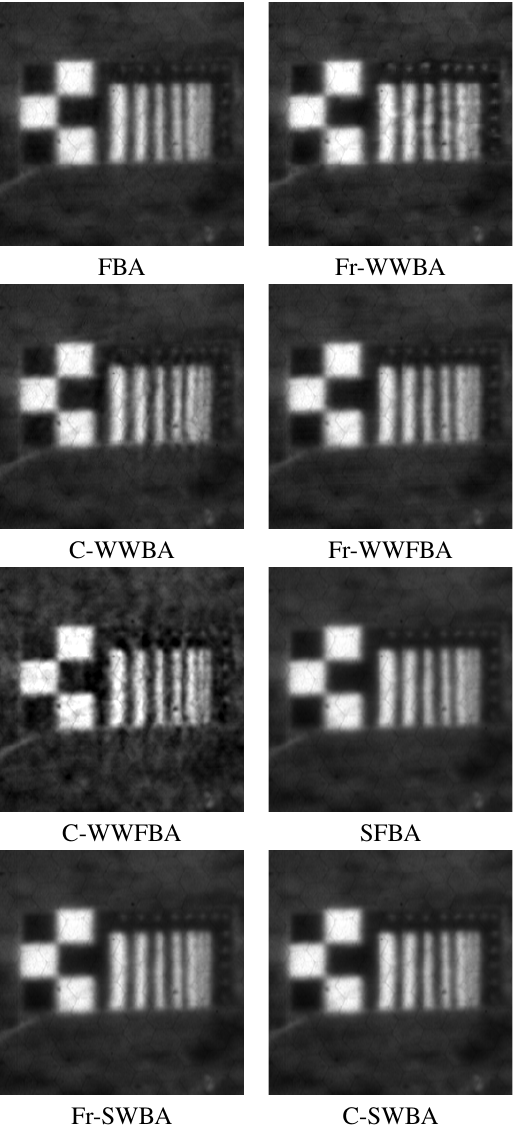} 
\end{center}
\caption{Restoration results on Barchart 1 sequence (50 frames) without non-rigid registration and $p=11$ (without any final denoising or sharpening step).}
\label{fig:nobar1}
\end{figure}

\begin{figure}[!t]
\begin{center}
\includegraphics[width=0.6\textwidth]{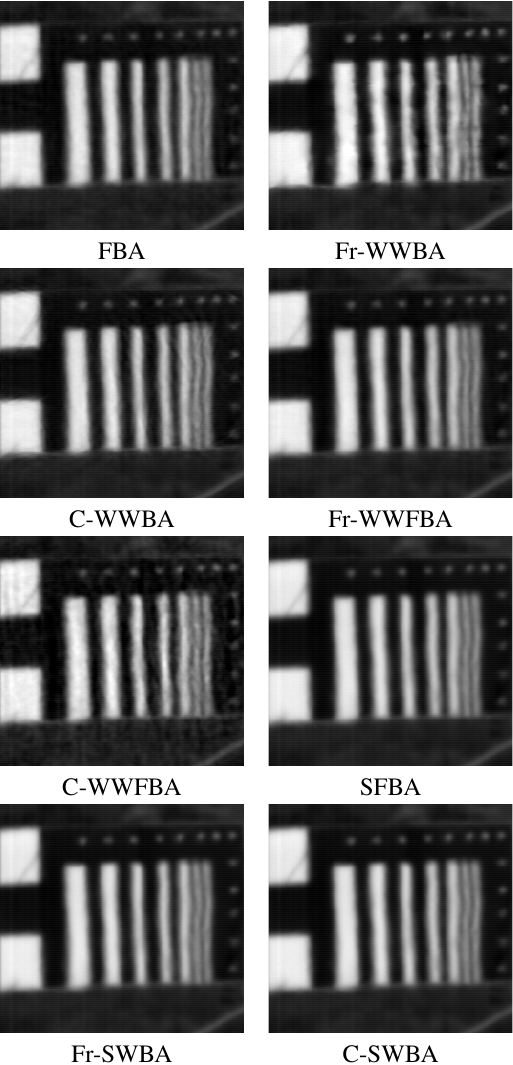} 
\end{center}
\caption{Restoration results on Barchart 2 sequence (50 frames) without non-rigid registration and $p=11$ (without any final denoising or sharpening step).}
\label{fig:nobar2}
\end{figure}

\begin{figure}[!t]
\begin{center}
\includegraphics[width=0.7\textwidth]{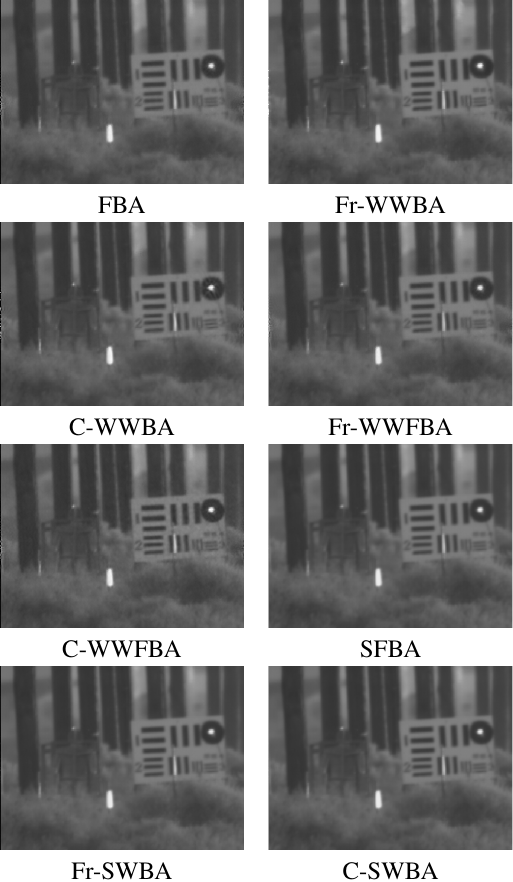} 
\end{center}
\caption{Restoration results on Barchart 3 sequence (50 frames) without non-rigid registration and $p=11$ (without any final denoising or sharpening step).}
\label{fig:nobar3}
\end{figure}

\begin{figure}[!t]
\begin{center}
\includegraphics[width=0.55\textwidth]{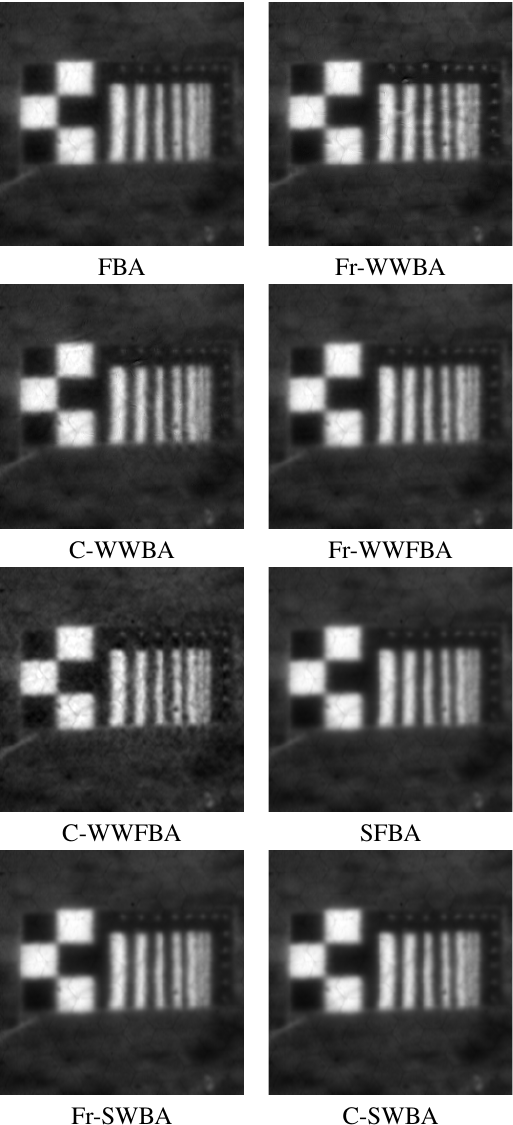} 
\end{center}
\caption{Restoration results on Barchart 1 sequence (50 frames) with non-rigid registration and $p=11$ (without any final denoising or sharpening step).}
\label{fig:withbar1}
\end{figure}

\begin{figure}[!t]
\begin{center}
\includegraphics[width=0.6\textwidth]{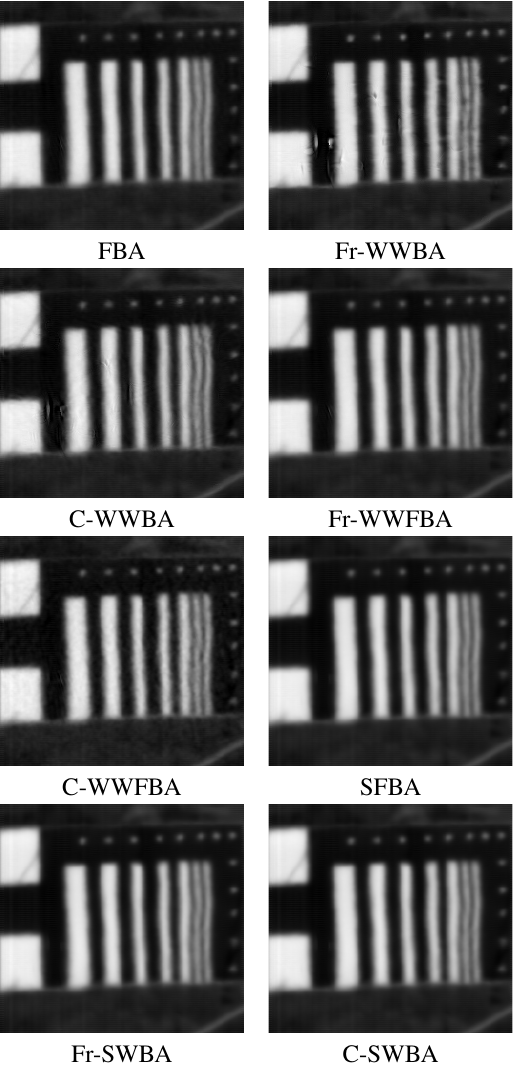} 
\end{center}
\caption{Restoration results on Barchart 2 sequence (50 frames) with non-rigid registration and $p=11$ (without any final denoising or sharpening step).}
\label{fig:withbar2}
\end{figure}

\begin{figure}[!t]
\begin{center}
\includegraphics[width=0.7\textwidth]{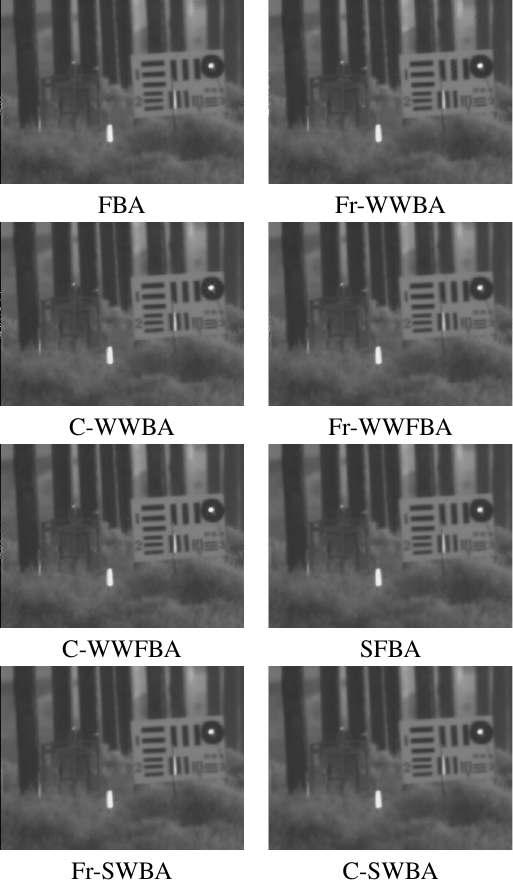} 
\end{center}
\caption{Restoration results on Barchart 3 sequence (50 frames) with non-rigid registration and $p=11$ (without any final denoising or sharpening step).}
\label{fig:withbar3}
\end{figure}

\begin{figure}[!t]
\begin{center}
\includegraphics[width=0.55\textwidth]{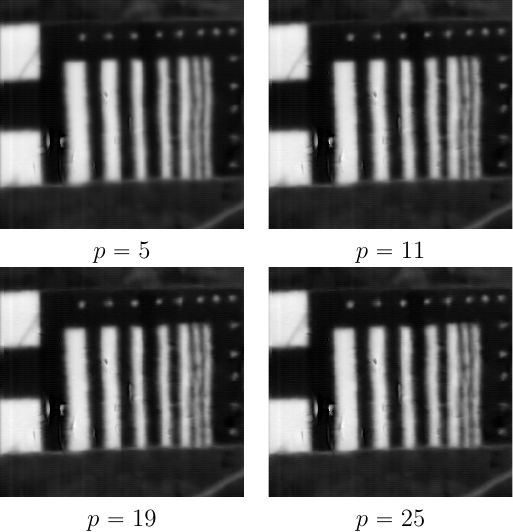} 
\end{center}
\caption{Influence of $p$ in the weighted methods. Here is it illustrated on Fr-WWBA with non-rigid registration.}
\label{fig:p}
\end{figure}

\begin{figure}[!t]
\begin{center}
\includegraphics[width=0.55\textwidth]{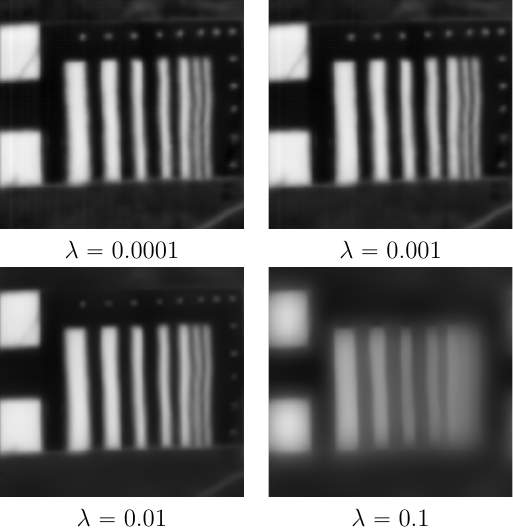} 
\end{center}
\caption{Influence of $\lambda$ in the sparse methods. Here is it illustrated on C-SWBA with non-rigid registration.}
\label{fig:lambda}
\end{figure}

In our experiments, we decided to use two popular families of wavelets in image processing: framelets and curvelets. Framelets are basically constructed in the same philosophy as for classic 2D tensor wavelets except that no downsampling is involved in the process. This particularity permits to guarantee translation invariant decompositions which is important for image processing. Moreover, it well known that such family form a tight frame which ensures easy reconstructions and permits to have algorithms which are less sensitive to important loss of information (see \cite{Selesnick} for more details). Curvelets also form a tight frame but their main particularity is in the fact that they also capture directional information hence providing better representations of geometric structures in images (see \cite{Candes1999a,Candes2005} for details).\\
In order to distinguish between the different algorithms, we will use the following acronyms:
\begin{itemize}
\item FBA: Fourier Burst Accumulation,
\item Fr-WWBA: Framelet based Weighted Wavelet Burst Accumulation,
\item C-WWBA: Curvelet based Weighted Wavelet Burst Accumulation,
\item Fr-WWFBA: Framelet based Weighted Wavelet Fourier Burst Accumulation,
\item C-WWFBA: Curvelet based Weighted Wavelet Fourier Burst Accumulation,
\item SFBA: Sparse Fourier Burst Accumulation,
\item Fr-SWBA: Framelet based Sparse Wavelet Burst Accumulation,
\item C-SWBA: Curvelet based Sparse Wavelet Burst Accumulation.
\end{itemize}
We experiment the proposed algorithms on three sequences denoted Barchart 1, Barchart 2 and Barchart 3 (frames 1 and 25 for each sequence are illustrated in Figure~\ref{fig:barchart}). These sequences were acquired with different equipments and meteorological (turbulence) conditions. We ran all algorithms (FBA, Fr-WWBA, C-WWBA, Fr-WWFBA, C-WWFBA, SFBA, Fr-SWBA and C-SWBA) on each sequence without any registration and with non-rigid registration. As suggested by the authors of \cite{Delbracio2015}, we fix $p=11$ for all weighted methods. For the 
sparse based methods, the parameter $\lambda$ is set to $\lambda=0.5$ for SFBA and $\lambda=0.001$ for Fr-SWBA and C-SWBA (the influence of these parameters will be discussed below). We used sequences of 50 frames in each test.\\
Figures~\ref{fig:nobar1}, \ref{fig:nobar2} and \ref{fig:nobar3} present the results obtained without any registration step. If, compared to the original frames, cleaner images are obtained, 
it is worth to notice that some geometric distortions remain in the reconstructed images thus comforting the idea that a non-rigid registration step is needed to deal with the 
turbulence problem.\\
The results obtained after using the non-rigid registration step are given in Figures~\ref{fig:withbar1}, \ref{fig:withbar2} and \ref{fig:withbar3}. It is clear that the non-rigid registration step is essential to correct the geometric distortions induced by the turbulence. For instance, we can notice that vertical bars in the different barcharts are straighter and sharper than in either the original and restored without non-rigid registration images. Comparing the different methods with non-rigid registration, the wavelet based options give clearer restored images than the FBA technique. The advantage looks like to be in favor of the sparse options as they provide images with more contrast than the weighted approaches.\\
As mentioned above, for the weighted algorithms, we used $p=11$ as proposed by the authors of \cite{Delbracio2015}. However, we tried several values of $p$ on the weighted wavelet burst methods to see the influence of $p$. Figure~\ref{fig:p} illustrates the obtained results and if the choice of $p$ does not seem very influential, we actually observe that $p=11$ seems to be a good trade-off between improving the image contrast and keeping only the most dominant coefficients. Unfortunately, no theoretical results support this value for $p$ and its investigation is a difficult task as it corresponds to a nonlinear optimization problem.\\
In Figure~\ref{fig:lambda}, we illustrate the influence of $\lambda$ on the C-SWBA algorithm, we verified that the same behavior also occurs for the Fr-SWBA method. It is clear that the bigger $\lambda$, the more regularized is the restored image. This behavior is easy to understand since when $\lambda$ increases, it corresponds to remove larger coefficients in the transform domain, i.e removing more details in the image. In all our experiments, we found that $\lambda=0.001$ for wavelet based methods and $\lambda=0.5$ for sparse Fourier burst are good trade-offs providing sufficient regularization while keeping details in the image.\\
In Table~\ref{tab:time}, we provide the different running times for each algorithms. All codes were run on a 2Ghz Inter(R) Xeon(R) E5-2640v2 (only a single core was used). All codes were implemented in Matlab 2015b and it is worth to notice that the used framelet library is not optimized and based on pure Matlab code (i.e does not have pre-compiled functions) while the libraries implementing the FFT and the curvelet transform were based on compiled code. This difference explains why the framelet based algorithms perform much slower than the other methods. We can notice, among all wavelet approaches, that the sparse methods are faster than the weighted based approaches.

\begin{table}[!t]
\begin{center}
\begin{tabular}{|c|c|c|c|} \hline
& Barchart 1 & Barchart 2 & Barchart 3\\ \hline
FBA & 2.29 & 2.08 & 2.27 \\ \hline
Fr-WWBA & 91.71 & 84.98 & 94.82 \\ \hline
C-WWBA & 27.59 & 18.46 & 19.51 \\ \hline
Fr-WWFBA & 97.97 & 92.91 & 103.37 \\ \hline
C-WWFBA & 9.8 & 9.39 & 11 \\ \hline \hline
SFBA & 1.63 & 1.74 & 1.91 \\ \hline
Fr-SWBA & 24.75 & 23.30 & 26.07 \\ \hline
C-SWBA & 4.35 & 4.05 & 4.77 \\ \hline
\end{tabular}
\end{center}
\caption{Running times (in seconds) for each algorithms.}\label{tab:time}
\end{table}

\section{Conclusion}\label{sec:conc}
In this paper, we extend the work of Delbracio et al. on restoring a static image from a sequence of blurred frames. We replace the rigid registration step by a non-rigid registration in order to be able to deal with geometric distorted frames. We also propose to use some wavelet domain instead of the Fourier domain and design two main approaches. Moreover, we suggest to replace the weighting process by a sparsity constraint in the considered domains. We show several experiments on real turbulence sequences to illustrate the efficiency of all described methods. These results permit to observe that the sparse based methods give better reconstructions and are faster than the weighted approaches. Future investigation can be made in terms of studying the best ``operators'' that could be used to combine the different frames in both the Fourier or wavelet domains. A more challenging theoretical question which should be addressed is why values of $p$ close to 11 seem to be the ``optimal'' ones?

\acknowledgments 
The authors want to thank the members of the NATO SET156 (ex-SET072) Task Group for the opportunity of using the data (sequences Barchart 1 and Barchart 2) collected during 
the 2005 New Mexico's field trials, and the Naval Air Warfare Center at China Lake, CA for providing the sequence Barchart 3.
This work is supported by the Air Force Office of Scientific Research grant FA9550-15-1-0065.

\clearpage
\bibliography{waveburst}
\bibliographystyle{spiejour}


\vspace{2ex}\noindent\textbf{J\'er\^ome Gilles} a Ph.D. degree in mathematics from the ENS Cachan, in 2006. He worked on signal and image processing with the French Ministry of Defense, from 2001 to 2010, and was Assistant Adjunct Professor with UCLA from 2010 to 2014, before he joined the Department of Mathematics at SDSU as an Assistant Professor. His research interests are applied harmonic and functional analysis for signal and image processing.

\vspace{2ex}\noindent\textbf{Stanley Osher} is Professor with UCLA's Department of Mathematics, Director of Special Projects at IPAM and has a joint faculty appointment with UCLA’s Electrical Engineering and Computer Science Departments. He was awarded several honors notably the 2012 AMS Fellow, the 2013 John von Neumann SIAM Lecture and the 2014 Gauss Prize. His main interest are numerical methods for image processing, PDEs and compressive sensing.

\end{document}